# Admiring the Great Mountain: A Celebration Special Issue in Honor of Stephen Grossberg's 80th Birthday

**Abstract**

This editorial summarizes selected key contributions of Prof. Stephen Grossberg and describes the papers in this 80th birthday special issue in his honor. His productivity, creativity, and vision would each be enough to mark a scientist of the first caliber. In combination, they have resulted in contributions that have changed the entire discipline of neural networks. Grossberg has been tremendously influential in engineering, dynamical systems, and artificial intelligence as well. Indeed, he has been one of the most important mentors and role models in my career, and has done so with extraordinary generosity and encouragement. All authors in this special issue have taken great pleasure in hereby commemorating his extraordinary career and contributions.

**Key contributions of Stephen Grossberg**
In brief, Grossberg as a biological neural modeler stands without peer, particularly in view of the predictive power of his models, and of their efficacy in unsupervised learning. His work has been independently touted (Hestenes, 1983) as the theoretical harbinger of a revolution in brain science, and subsequent confirmation by experimental psychologists of these models has borne out this assessment. It has been a constant marvel that mathematics works so well to model the many subtleties of scientific phenomena. Grossberg's work is as strong an example of this principle as any.

He introduced, and has done more than anyone to develop, one of the most important computational paradigms ever; autonomous, self-correcting, biological intelligence. This includes but goes beyond unsupervised learning. The paradigm explains how individual humans or animals can learn to autonomously adapt in real time to complex and changing environments that are filled with unexpected events.

This work began in 1957, when Grossberg was still a freshman in college. Motivated by clearly-open questions in his Psychology coursework, Grossberg introduced the computational paradigm for linking brain mechanisms to psychological functions using real-time nonlinear neural networks. As part of this breakthrough, he derived classical laws for short-term memory, medium-term memory, and long-term memory that are still used, in some variant, by essentially all biological neural network modelers today.

Over the next six decades, Grossberg made a continuous stream of ground-breaking, and often revolutionary, theoretical breakthroughs continuing to the present. There are far too many to catalog in one or even several reviews; in fact, several books have been published on the subject.

I will therefore focus on just three of my favorite themes in his research.

1. His fundamental breakthrough for which he is best known was the introduction of Adaptive Resonance Theory (ART) (Grossberg, 1976). Grossberg and his colleagues have continually developed ART to its present status as the most advanced cognitive and neural theory about how humans and other animals learn to attend, recognize, and predict objects and events in a changing world. ART presently enjoys an unrivaled explanatory range. It has explained the data from many hundreds of psychological and neurobiological experiments, and scores of its predictions have been confirmed by psychological and neurobiological data, including *all* of the predictions about the theory's foundational mechanisms. The unique combination of learning, memory, and prediction properties of ART is equally important, with far-reaching consequences. ART explains how to carry out arbitrary combinations of both unsupervised and supervised learning. The learning of sequences of events is incremental, fast or slow—fast learning can even learn a database in one learning trial—and automatically adjustable to learn both concrete and abstract information in response to arbitrarily large non-stationary databases. The ability to adjust category generality to match the statistics of a particular database was invented by Grossberg as part of ART and is called *vigilance control.* Grossberg has described the brain mechanisms of vigilance control in great detail, including their specific anatomical, neurophysiological and biochemical substrates (Grossberg & Versace, 2008) and has shown how, when vigilance control breaks down in specific ways, it can lead to symptoms of mental disorders such as Alzheimer's disease, autism, medial temporal amnesia, and abnormal slow wave sleep (Grossberg, 2017a). After learning occurs in ART, it enables direct access to the globally best matching learned category in its repertoire, without any memory search, no matter how many additional categories are learned subsequently. Otherwise expressed, ART solves the global minimum problem. Even more fundamentally, ART solves what Grossberg called the *stability-plasticity dilemma.* In systems that solve this problem, learned memories are dynamically self-stabilizing to prevent unpredictable forgetting of previously learned, but still useful, information in response to future learning. Without such a guarantee against catastrophic forgetting, a learning algorithm is unreliable. It is therefore significant that other machine learning algorithms are well known to suffer catastrophic forgetting. To help place these results in context, progress since then is well-summarized in (Grossberg, 2013) and (Grossberg, 2017b). I'll add a personal favorite; the discussion in (Grossberg, 1980) is a superb interplay between his bold but clear thought experiments, the psychological principles he was elucidating, and the mathematical formulations needed for the tools he was developing. It is essential to emphasize that ART is not a neural network architecture or even a collection of them. Instead, *ART is a learning theory. Succinctly summarized, it states that, for neural networks that use feedback, that is to say, for the most powerful ones, learning is regulated by resonance.* For many people who appreciate the beauty and importance of resonance in feedback systems, reading this theory for the first time is a vicarious Eureka moment. Digging deeper is certain to increase one's appreciation even further. Much evidence for this theory has been gathered (and documented in the most recent papers cited above,) and its applied efficacy is also well-demonstrated, as elucidated in the engineering-oriented review article (D. C. Wunsch, 2009) and in an updated article at the end of this issue, discussed later in this editorial. This is an appropriate segue to the many neural network architectures based on ART, described below.

2. As compelling as ART is theoretically, for solving practical problems the many related neural networks architectures have been useful to many researchers. The first of these (Carpenter & Grossberg, 1987), often referred to as ART1, is his most highly-cited paper to date. This breakthrough article proves theorems about the learning, recognition, and prediction properties of ART1, an algorithm that has been used in many large-scale applications to technology, including a parts retrieval system that was used for design retrieval of structures in the Boeing 777 (Caudell, Smith, Johnson, Wunsch, & Escobedo, 1992; Caudell, Smith, Johnson, & Wunsch, 1991; Smith, Escobedo, Anderson, & Caudell, 1997). Grossberg and his collaborators have developed many other architectures. Key contributions to these architectures (within this one major facet of his research) are Fuzzy ART (Carpenter, Grossberg, & Rosen, 1991), ARTMAP (Carpenter, Grossberg, & Reynolds, 1991), and Fuzzy ARTMAP (Carpenter, Grossberg, Markuzon, Reynolds, & Rosen, 1992) – although these emphasize neural network models with very general capabilities. Grossberg has also contributed many ART-based neural network models that fit more specific applications, as well as models not relying on ART. From his website (Grossberg, 2019), he develops "brain models of vision and visual object recognition; audition, speech, and language; development; attentive learning and memory; cognitive information processing and social cognition; reinforcement learning and motivation; cognitive-emotional interactions; navigation; sensory-motor control and robotics; and mental disorders. These models involve many parts of the brain, ranging from perception to action, and multiple levels of brain organization, ranging from individual spikes and their synchronization to cognition." (The website goes on to characterize even more areas of seminal contributions, substantiating all of them.) While I've read and admire many of these, I can't do justice to them within the scope of this editorial. Building just on the most general class of ART architectures, there are numerous other examples, including independent contributions from various researchers, notably LAPART (Healy, Caudell, & Smith, 1993), Fuzzy Min-Max ART (Carpenter et al., 1992), TD-FALCON (Tan, Lu, & Xiao, 2008), TopoART (Tscherepanow, 2010), Dual-Vigilance ART (Brito da Silva, Elnabarawy, & Wunsch, 2019), Distributed Dual-Vigilance ART, and for biclustering, BARTMAP (Xu & Wunsch, 2011). The point is that Grossberg's insights have inspired many neural network models by himself and a large number of other researchers – evidence of the extraordinary impact of his work.

3. Just as important as the proliferation of neural network designs built upon ART are the mathematical foundations of Grossberg's theories and the resulting neural architectures. Grossberg used the aforementioned thought experiment approach to develop mathematical models that were both original and far-reaching. The reason these models were so fundamental is that he typically started with a question of the form, "What is the minimal model that would be capable of …" I'll constrain my remarks to a few favorites;

there are many more. An early example, regarding Pavlovian conditioning, appeared in the Proceedings of the National Academy of Sciences (Grossberg, 1971). He made arguments about why certain physiological conditions created necessary mathematical assumptions, resulting in elegant theorems for simple but powerful mathematical models. This has been his method before and since. He published a seminal analysis of feedforward and feedback, cooperative-competitive dynamical systems as early as 1973, but my favorite summary is in chapter 2 of (Grossberg, Stephen; Kuperstein, 1986). It covers a tremendous amount of material, and a solid understanding of it is necessary for a much more efficient design of adaptive or hardwired neural networks (e.g. (Cai, Prokhorov, & Wunsch, 2007; Wunsch, 2000) and citing papers.) As Grossberg's mathematical analyses developed further, he continued to ask simple questions while developing new methods to answer them. The most famous of these, (Cohen & Grossberg, 1983), (often referred to by others as the Cohen-Grossberg theorem) introduces a general class of nonlinear dynamical systems for which a Lyapunov function is defined as part of the proof of global convergence to possibly infinitely many equilibrium points. These systems are interpreted as content addressable memories in a broad range of recurrent neural network architectures. The 1983 article builds on a few of Grossberg's earlier works, my favorite of which is (Grossberg, 1978), which is deeply intricate, stunningly creative, and broad-ranging in its implications. This article introduces a large class of competitive dynamical systems and introduces a highly original Method of Jumps to embed a discrete decision scheme into it, thereby making rigorous the intuition that one way to understand a competition is to keep track of who is winning it through time. Each Jump is a discrete decision that occurs when a new population begins to win. Grossberg used his Method of Jumps to prove global theorems about both oscillatory and convergence dynamics in competitive systems. Persistent oscillations tend to occur when there are Jump cycles, and convergence tends to occur when there are not. The article focuses on a general class of competitive systems that always converge to one of possibly infinitely many equilibrium points. It is prescient in that it explicitly describes implications not only for neural networks, but also economic and social networks, decades before most people were even thinking about such topics! One intriguing economic application proves sufficient conditions that guarantee that a competitive market's price will be stable, while every firm can track publicly known market prices to make its own decisions about production and savings, with no knowledge about the policies of the possibly arbitrarily many other firms with which it is competing. The theorem provides, in other words, a mathematical description of how Adam Smith's Invisible Hand (Smith, 1776) can work in a competitive market.

As impressive as these accomplishments of Grossberg are, they are augmented by his accomplishments in technological leadership of the highest order via his development of the profession, including his pivotal role in the formation of the International Neural Networks Society and the journal *Neural Networks*.

I hope that Grossberg's genius will continue to illuminate science with his continued research for many years to come. His visionary discoveries and tireless leadership have already created an extraordinary legacy. On the occasion of his 80th birthday it is a delight to honor his leading research on introducing, developing, and mathematically characterizing processes of biological learning.

**This Issue**

The articles in this special issue are a glimpse into how Grossberg's research has influenced others. We begin with a paper by his closest collaborator and spouse. (Carpenter, 2019) discusses using Self-supervised ART for continual learning in unpredictable environments, placing this discussion into the context of modern challenges in Artificial Intelligence. (Kosko, 2019) shows that noise injection benefits the performance of (ART-inspired) bidirectional backpropagation on classifiers as well as generative adversarial networks. (Seiffertt, 2019) extends his previous seminal contributions relating time-scales calculus (Bohner & Peterson, 2001) to reinforcement learning (Seiffertt, Sanyal, & Wunsch, 2008) and backpropagation (Seiffertt & Wunsch, 2010) to now relate this important generalization between continuous and discrete time for the first time to ART. Those who use machine learning with both continuous and discrete time signals would do well to study this paper and its references. (Healy & Caudell, 2019) use category theory to develop a representation of hierarchical episodic memory building on various important memory models including Grossberg's early work. ART has long been considered a suitable architecture for sensor fusion. (Tan, Subagdja, Wang, & Meng, 2019) makes the case that Fusion ART can go beyond sensor fusion by combining several information channels, resulting in more complex tasks such as combining multiple learning modalities. (Levine, 2019) explore the consistency of decision making that goes beyond economic utility maximization to consider the broader range of behaviors encountered in real life. (Zeid & Bullock, 2019) describes models and evidence for complex sequence learning, such as musical sequences. (Patel, Hazan, Saunders, Siegelmann, & Kozma, 2019) converts deep Q-learning to a spiking neural networks model, showing improved robustness in the process. (Wandeto & Dresp-Langley, 2019) explores how quantization error in neural networks can reliably discriminate fine differences in local contrast. (Brna et al., 2019) uses modulation of uncertainty in an ART context to enable Lifelong Learning, including self-supervised and one-shot learning. (Meng, Tan, & Miao, 2019) introduces Salient-Aware ART, which builds on his previous works (Meng, Tan, & Wunsch, 2016; Meng, Tan, & Wunsch II, 2019) that customize update rules to each ART cluster; while retaining linear computational complexity. This is useful for the important problem of clustering sparse data. (Pessoa, 2019) makes the case for investigating the dynamic, multivariate structure of brain data to understand emotion and cognition. Finally, (Brito da Silva, Elnabarawy, & Wunsch II, 2019) provides the most comprehensive survey yet on various modifications to ART for engineering applications. All the special issue authors are pleased to offer this enduring commemoration of Grossberg's extraordinary career.

In addition to the valuable contributions of each author in this issue, discussions of an early version of this editorial with Mario Aguilar-Simon, Jose L. Contreras-Vidal, Morris W. Hirsch, Daniel S. Levine and Hava T. Siegelmann are gratefully acknowledged. The author would also like to thank Stephen Grossberg and Gail Carpenter for their incredible intellectual and personal hospitality and encouragement.